\documentclass[runningheads,a4paper]{llncs}

\usepackage[utf8]{inputenc}
\usepackage{amsmath,amssymb,mathtools}
\usepackage[detect-weight=true, binary-units=true]{siunitx}
\usepackage{url,hyperref,cleveref}
\Crefname{equation}{Eq.}{Eqs.}
\Crefname{figure}{Fig.}{Figs.}
\Crefname{tabular}{Tab.}{Tabs.}
\Crefname{section}{Sec.}{Secs.}
\usepackage[linesnumbered, ruled, vlined]{algorithm2e}
\crefname{algocf}{alg.}{algs.}
\Crefname{algocf}{Algorithm}{Algorithms}
\usepackage[normalem]{ulem} 
\usepackage{color,colortbl,etoolbox} 
\usepackage[dvipsnames]{xcolor}
\robustify\bfseries
\robustify\uline
\usepackage{comment}
\usepackage{graphicx}
\usepackage{multirow,booktabs,makecell,array}
\usepackage[caption=false]{subfig}
\usepackage{csquotes} 
\usepackage[inline]{enumitem}
\usepackage{hhline} 
\usepackage{array,multirow}
\spnewtheorem{prop}{Proposition}{\bfseries}{\itshape}
\spnewtheorem{thm}{Theorem}{\bfseries}{\itshape}

\definecolor{LightGray}{gray}{0.9}

\begin{document}

\title{Genetic Programming is Naturally Suited to Evolve Bagging Ensembles}

\author{
    Marco Virgolin\orcidID{0000-0001-8905-9313}
}
\institute{
    Chalmers University of Technology, Gothenburg, Sweden\\ \email{marco.virgolin@chalmers.se} 
}
\authorrunning{M.~Virgolin}

\maketitle

\begin{abstract}
    Learning ensembles by bagging can substantially improve the generalization performance of low-bias, high-variance estimators, including those evolved by Genetic Programming (GP). 
  To be efficient, modern GP algorithms for evolving (bagging) ensembles typically rely on several (often inter-connected) mechanisms and respective hyper-parameters, ultimately compromising ease of use.
  In this paper, we provide experimental evidence that such complexity might not be warranted.
  We show that minor changes to fitness evaluation and selection are sufficient to make a simple and otherwise-traditional GP algorithm evolve ensembles efficiently.
  The key to our proposal is to exploit the way bagging works to compute, for each individual in the population, multiple fitness values (instead of one) at a cost that is only marginally higher than the one of a normal fitness evaluation.
  Experimental comparisons on classification and regression tasks taken and reproduced from prior studies show that our algorithm fares very well against state-of-the-art ensemble and non-ensemble GP algorithms.
  We further provide insights into the proposed approach by (i) scaling the ensemble size, (ii) ablating the changes to selection, (iii) observing the evolvability induced by traditional subtree variation.
  Code: \url{https://github.com/marcovirgolin/2SEGP}.
\end{abstract}

\section{Introduction}
Learning ensembles by \emph{bagging}, i.e., aggregating the predictions of low-bias, high-variance estimators fitted on different samples of the training set (see \Cref{sec:ensemble-learning-setting} for a more detailed description), can improve generalization performance significantly~\cite{bishop2006pattern,breiman1996bagging,dietterich2000ensemble,opitz1999popular}.
Random forests are a remarkable example of this~\cite{biau2012analysis,breiman2001random}. 
At the same time, mixed results have been found when using deep neural networks~\cite{chen2010multiobjective,fort2019deep,lakshminarayanan2017simple,strauss2017ensemble}.
For Genetic Programming (GP)~\cite{koza1992genetic,poli2008field}, bagging has generally been found to be beneficial~\cite{dick2018evolving,iba1999bagging,keijzer2000genetic,rodrigues2020ensemble}.
Since in a \emph{classic} GP algorithm the outcome of the evolution is \emph{one} best-found individual (i.e., the estimator that best fits the training set), perhaps the simplest way to build an ensemble of GP individuals is to evolve multiple populations independently, and aggregate the outcomes.
However, since a GP population can naturally host diverse individuals, it makes sense to seek ways to evolve the ensemble in one go and save substantial computational resources.

Many ensemble learning GP-based approaches have been proposed so far (see \Cref{sec:related} for an overview). 
We can broadly categorize them in two classes: \emph{Simple Independent Ensemble Learning Approaches} (SIEL-Apps), and \emph{Complex Simultaneous Ensemble Learning Algorithms} (CSEL-Algs).
SIEL-Apps form an ensemble of estimators by repeating the execution of a (typically classic) GP algorithm that produces, each time, a single estimator.
As said before, this idea is simple but inefficient.
Instead, CSEL-Algs make use of a number of novel mechanisms and respective hyper-parameters to obtain an ensemble in one go.
For this reason, CSEL-Algs can be very efficient, but also quite complex and thus difficult to adopt in practical applications.
Moreover, from a scientific standpoint, it may be hard to assess which moving-parts of a CSEL-Alg are really needed and which are not.

In this paper, we seek to obtain the best of both worlds: A GP algorithm that learns ensembles efficiently (e.g., without repeating multiple evolutions) and is simple enough to be thought as a possible minimal/natural extension of classic GP.
Specifically, given a classic tree-based GP algorithm, we introduce only (arguably) minor modifications to fitness evaluation and selection, with the goal of \textbf{making the population specialize uniformly across different realizations of the training set} (in the context of bagging, these are called \emph{bootstrap samples}, see \Cref{sec:ensemble-learning-setting}).
The proposed modifications are time-efficient.
We call the resulting algorithm \emph{Simple Simultaneous Ensemble Genetic Programming} (2SEGP) and show that, despite its simplicity, 2SEGP is competitive with State-of-the-Art (SotA) ensemble and non-ensemble GP algorithms.
We do this by reporting and reproducing results from recent literature on real-world benchmark classification and regression datasets.
Moreover, to better understand what matters when learning bagging ensembles in GP, we include experiments that dissect our algorithm.



\section{Related work}
\label{sec:related}
In this paper we focus on ensemble learning intended as bagging (see \Cref{sec:ensemble-learning-setting}), when GP is used to learn (evolve) the estimators.
We do not consider ensemble learning intended as boosting, i.e., the iterative fitting of weak estimators to (weighted) residuals~\cite{chen2016xgboost,friedman2000additive,friedman2001greedy}. 
For the reader interested in boosting GP, we refer, e.g., to~\cite{folino2007mining,iba1999bagging,paris2001applying,ruberto2020sgp,de2010applying}.
Similarly, we do not consider works where, even if GP was used to decide how to aggregate the estimators, these were learned by other (machine learning) algorithms than GP~\cite{bi2019automated,bi2020genetic,johansson2006genetically,langdon2001genetic,langdon2002combining}.

Starting with SIEL-Apps, we remark that the works in this category mostly focus on \emph{how the aggregation of GP-evolved estimators can be improved}, rather than on how to evolve ensembles efficiently.
For example, some early works look into improving the ensemble prediction quality by weighing member predictions by a measure of confidence~\cite{iba1999bagging} or by bypassing outlier member predictions~\cite{keijzer2000genetic}.
Further investigations have been carried out across problems of different nature, in~\cite{imamura2001fault,imamura2002n,sobania2018covsel}. An SIEL-App is also used in~\cite{veeramachaneni2015flexgp}, yet this time with a non-classic GP algorithm where individuals are linear combinations of multiple trees and the evolution is made scalable by leveraging on-line, parallel computing.
Other works in this category are~\cite{kotanchek2008trustable,tran2018genetic,zhang2004genetic}, respectively for hybridization with multi-objective evolution, incomplete data, and large-scale data.

CSEL-Algs are of most interest w.r.t. the present work as they attempt to evolve an ensemble in an efficient manner.
In~\cite{bhowan2012evolving}, e.g., multi-objective GP is used to build ensembles where the members are Pareto non-dominated individuals. Importantly, having multiple objectives is a prerequisite for this proposal (not the case here).
Multifactorial GP is used in~\cite{wen2016learning} to evolve ensembles of decision tree-like individuals that each interpret the dataset features differently.
More recently,~\cite{wang2019novel} proposed the Diverse Niching Genetic Programming (DivNichGP) algorithm, which works in single-objective and manages to obtain an ensemble by maintaining population diversity by (i) Using bootstrap sampling every generation to constantly vary the training data distribution, and (ii) Including a niching mechanism. 
Niching is further used at termination in order to pick the final ensemble members from the population, and requires two dedicated hyper-parameters to be set.
Another recent investigation is~\cite{dick2018evolving}, where ensembles are learned to reduce the typical susceptibility of symbolic regression GP algorithms to outliers. 
In that work, spatially-clustered individuals (e.g., as neighboring nodes of a toroidal graph) compete in fitting different bootstrap samples~\cite{tomassini2006spatially}.
This algorithm requires to choose the graph and cluster structure as well as the way computational resources should be distributed on the graph nodes.
Lastly, in~\cite{rodrigues2020ensemble} ensemble learning is realized by the simultaneous co-evolution of a population of estimators (trees), and a population of ensemble candidates (forests).
For this algorithm, alongside the hyper-parameters for the population of trees, one needs to set the hyper-parameters for the population of forests (e.g., for variation, selection, and voting method).

We remark that, in order to ameliorate for the complexity introduced in CSEL-Algs, the respective works provide recommendations on default hyper-parameter settings.
Even so, we believe that these algorithms can still be considered sufficiently complex that pursuing a simpler approach remains a worthwhile endeavour. 
We include the three CSEL-Algs from~\cite{dick2018evolving,rodrigues2020ensemble,wang2019novel} (among other GP algorithms) in our comparisons.

\section{Learning bagging ensembles by minor modifications to classic GP}\label{sec:method}
We now describe how, taken a classic GP algorithm that returns a single best-found estimator, one can evolve bagging ensembles. 
In other words, how to obtain 2SEGP from classic GP.
We assume the reader to be familiar with the workings of a classic tree-based GP algorithm, and refer, e.g., to~\cite{poli2008field} (Chapters 2--4).

The backbone of our proposal consists of two aspects:
\begin{enumerate*}[label=(\roman*)]
\item \emph{Evaluate a same individual according to different realizations of the training set (i.e., bootstrap samples)}; and
\item \emph{Let the population improve uniformly across these realizations}.
\end{enumerate*}
To achieve these aspects, we only modify fitness evaluation (we also describe the use of linear scaling~\cite{keijzer2003improving} as it is very useful in practice) and selection.
We do not make any changes to variation: Any parent can mate with any other (using classic subtree crossover), and any type of genetic material can be used to mutate any individual (using classic subtree mutation).
Our intuition is that exchanging genetic material between estimators that are best on different samples of the training set is not detrimental because these samples are themselves similar to one another (we provide insights about this in~\Cref{sec:improvements-variation}).

We proceed by recalling how bagging works, followed by describing the modifications we propose, for the sake of clarity, first to selection and then to fitness evaluation.

\subsection{Bagging}\label{sec:ensemble-learning-setting}
As aforementioned, we focus on learning ensembles by bagging, i.e., bootstrap aggregating~\cite{breiman1996bagging}.
We use traditional bootstrap, i.e., we obtain $\beta$ realizations of the training set $\mathbb{T}_1, \dots, \mathbb{T}_\beta$, each with as many observations as the original training set $\mathbb{T} = \{(\mathbf{x}_i, y_i)\}^n_{i=1}$, by uniformly sampling from $\mathbb{T}$ with replacement.
Aggregation of predictions is performed the traditional way, i.e., by majority voting (i.e., mode) for classification, and averaging for regression.
One run of our algorithm will produce an ensemble of $\beta$ members where each member is the best-found individual (i.e., elite) according to the fitness measured on the bootstrap sample $\mathbb{T}_j$, with $j=1, \dots, \beta$.

\subsection{Selection for uniform progress across the bootstrap samples}\label{sec:our-selection}
We employ a remarkably simple modification of truncation selection that is applied after the offspring population has been obtained by variation of the parent population, i.e., in a $(\mu + \lambda)$ fashion.
The main idea is to select individuals in such a way that progress is uniform across all the bootstrap samples $\mathbb{T}_1, \dots, \mathbb{T}_\beta$.
To this end, we now make the assumption that each individual does not have a single fitness value, rather, it has $\beta$ of them, one per bootstrap sample $\mathbb{T}_j$.
We show how these $\beta$ fitness values can be computed efficiently in \Cref{sec:fitness-eval}.

Pseudocode for the modified truncation selection is given in \Cref{alg:our-selection}. 
Very simply, we perform $\beta$ truncation selections, each focused on one of the $\beta$ fitness values, and where the $\left( n_\textit{pop} / \beta \right)$ top-ranking individuals are chosen each time.
Note that this selection ensures weakly monotonic fitness decrease across all the bootstrap samples. 
Note also that a same individual can obtain multiple copies if it has fitness values such that it is top-ranking according to multiple bootstrap samples.

Lastly, one can see that the computational complexity of this selection method is determined by sorting the population $\beta$ times and copying individuals, i.e., $\mathcal{O}(\beta n_\text{pop} \log n_\text{pop} + n_\text{pop} \ell)$, under the assumption that $\ell$ is the (worst-case) size of an individual (in the case of tree-based GP, the number of nodes). 
As we will show in \Cref{sec:fitness-eval} below, the cost of fitness evaluation over the entire population will dominate the cost of selection. 

We provide insights on the use of this selection method in~\Cref{sec:insight-selection}.

\begin{algorithm}
\SetKwData{Left}{left}\SetKwData{This}{this}\SetKwData{Up}{up}
\SetKwFunction{Union}{Union}\SetKwFunction{FindCompress}{FindCompress}
\SetKwInOut{Input}{input}\SetKwInOut{Output}{output}

\Input{$\mathbb{P}$ (parent pop.), $\mathbb{O}$ (offspring pop.), $\beta$ (ensemble size)}
\Output{$\mathbb{P}^\prime$ (new pop. of selected individuals)}
\BlankLine
     
 $\mathbb{Q} = \text{join}( \mathbb{P} , \mathbb{O} )$\;
 $\mathbb{P}^\prime \leftarrow [ ]$\;
 
 \For{$j \in 1, 2, \dots, \beta$}{
  sort $\mathbb{Q}$ according to the $j$th fitness value\;
  \For{$k \in 1, 2, \dots, \left( n_\text{pop} / \beta \right)$ }{
    $\mathbb{P}^\prime \leftarrow \text{join}( \mathbb{P}^\prime , [ \mathbb{Q}_k ] )$\;
  }
 }
 \Return $\mathbb{P'}$\;
 \caption{Our simple extension of truncation selection.}
 \label{alg:our-selection}
\end{algorithm}

\subsection{Fitness evaluation on all bootstrap samples}\label{sec:fitness-eval}
A typical fitness evaluation in GP comprises
\begin{enumerate*}[label=(\roman*)]
    \item\label{item:output}Computing the output of the individual in consideration;
    \item\label{item:loss}Computing the loss function between the output and the label.
\end{enumerate*}
Both steps are performed using the original training set $\mathbb{T} = \{(\mathbf{x}_i, y_i)\}^n_{i=1}$.
Recall that the computation cost of step~\ref{item:output} is $\mathcal{O}(\ell n)$, because we need to compute the $\ell$ operations that compose the individual for each observation in the training set.
Step~\ref{item:loss} takes $\mathcal{O}(n)$ but is additive, thus the total asymptotic cost ultimately amounts to $\mathcal{O}(\ell n)$.

Since we wish the population to evolve uniformly well across the bootstrap samples, our selection method needs each individual to have a fitness value for each bootstrap sample. 
In other words, we need to compute the fitness w.r.t. $\mathbb{T}_1, \dots, \mathbb{T}_\beta$.
A naive solution would be to repeat steps~\ref{item:output} and~\ref{item:loss} for each $\mathbb{T}_j$, leading to a time cost of $\mathcal{O}(\beta \ell n)$; Ultimately the same cost an SIEL-App would have (although distributed across multiple evolutions).

To improve upon the naive cost $\mathcal{O}(\beta \ell n)$, we make the following observation. In many machine learning algorithms, the specific realization of the training set determines the structure of the estimator that will be learned in an explicit (and possibly deterministic) way. 
For example, to learn a decision tree, the training set is used to determine what nodes are split and what condition is applied~\cite{breiman1984classification}. Consequently, when making bagging ensembles of decision trees (i.e., random forests~\cite{breiman2001random}), one needs to build each decision tree as a function of the respective $\mathbb{T}_j$, and so a multiplicative $\beta$ term in the asymptotics cannot be avoided.
The situation is different in GP. In GP, the structure of an individual emerges as an implicit byproduct of the whole evolutionary process; Fitness evaluation, in particular, is not responsible for altering structure. We exploit this.

Recall that each $\mathbb{T}_j$ is obtained by bootstrap of the original $\mathbb{T}$, thus contains only elements of $\mathbb{T}$. 
It follows that an individual's output computed over the observations of $\mathbb{T}_j$ contains only elements that are also elements of the output computed over $\mathbb{T}$.
So, if we compute the output over $\mathbb{T}$, we obtain the output elements for $\mathbb{T}_j, \forall j$.
Formally, let $\mathbb{S}_j$ be the collection of indices that identifies $\mathbb{T}_j$, i.e.,  $\mathbb{S}_j = [s^j_1, \dots, s^j_n]$ s.t.\ $s^j_l \in \{1, \dots, n\}$ and $\{(\mathbf{x}_k, y_k)\}_{k \in \mathbb{S}_j} = \{(\mathbf{x}_k, y_k)\}^{s^j_n}_{k=s^j_1} = \mathbb{T}_j$. 
Then one can:
\begin{enumerate}
    \item \label{proof:step-1} Compute \emph{once} the output of the estimator over $\mathbb{T}$, i.e.,  $\{ o_i\}^n_{i=1}$;
    \item \label{proof:step-2} For $j=1, \dots, \beta$, assemble a $\mathbb{T}_j$-specific output $\{ o_k\}_{k \in \mathbb{S}_j}$ from $\{o_i\}^n_{i=1}$;
    \item \label{proof:step-3} For $j=1, \dots, \beta$, compute $\textit{Loss}( \{y_k\}_{k \in \mathbb{S}_j},  \{ o_k \}_{k \in \mathbb{S}_j} )$ as $j$th fitness value.
\end{enumerate}
Step \ref{proof:step-1} costs $\mathcal{O}(\ell n)$, step \ref{proof:step-2} and step \ref{proof:step-3} cost $\mathcal{O}(\beta n)$, they are executed in sequence:
\begin{equation}\label{eq:fitness-cost}
    \mathcal{O}(\ell n) + \mathcal{O}(\beta n) + \mathcal{O}(\beta n) = \mathcal{O}( n(\ell + \beta)).
\end{equation}

\begin{figure}
    \centering
    \includegraphics[width=0.7\linewidth]{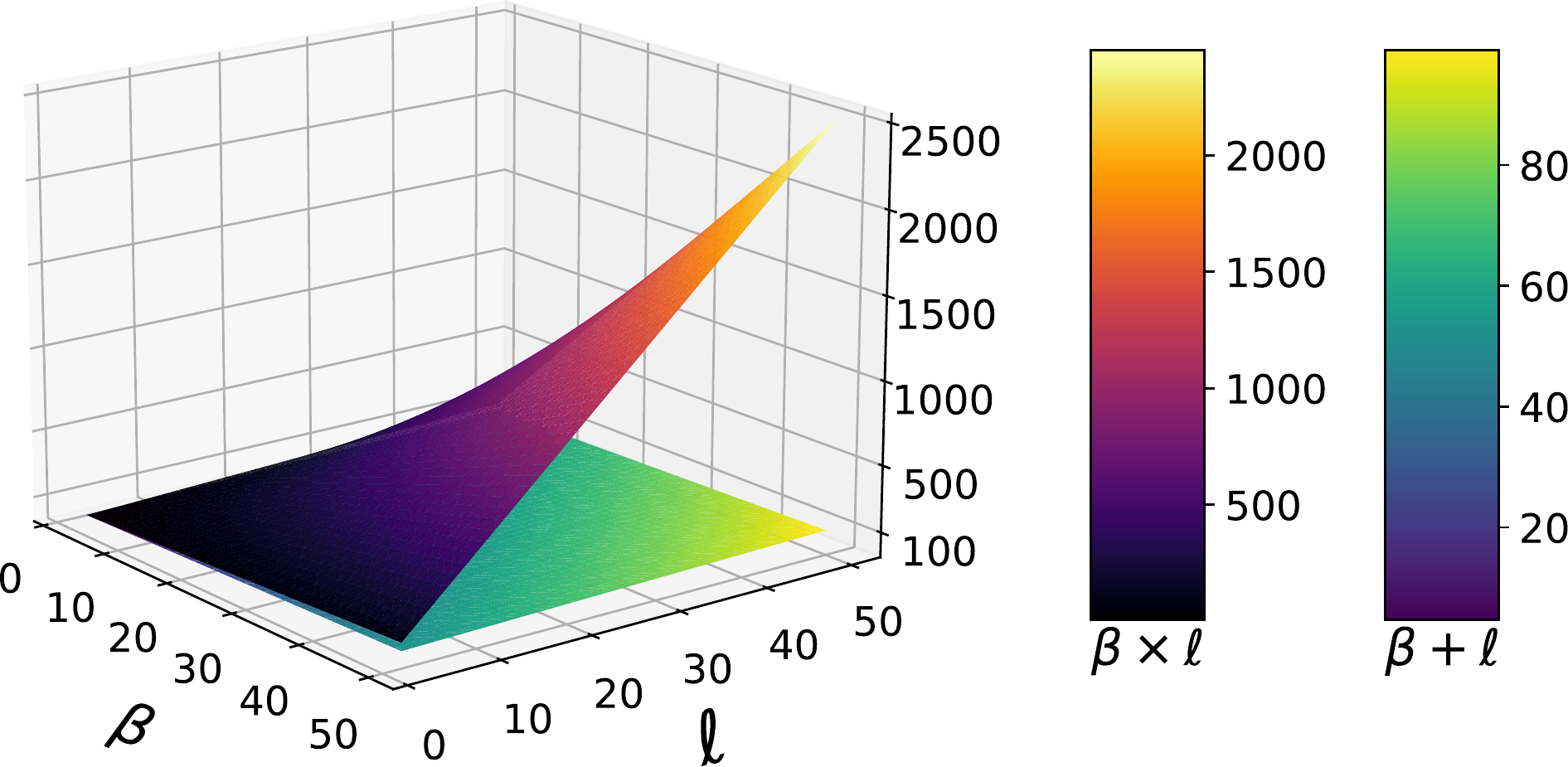}
    \caption{Scaling (vertical axis) of $\beta \times \ell$ and $\beta + \ell$.}
    \label{fig:naive-vs-smarter-bootstrap}
\end{figure}

This method is asymptotically faster than re-computing the output over each $\mathbb{T}_j$ whenever $\ell + \beta < \ell \beta$---basically in any meaningful scenario. 
\Cref{fig:naive-vs-smarter-bootstrap} shows at a glance, for growing $\beta$ and $\ell$, that the additive contribution $\beta + \ell$ quickly becomes orders of magnitudes better than the multiplicative one $\beta \times \ell$. Memory-wise, all we need is to store each $\mathbb{S}_j$ at initialization, which costs $\mathcal{O}(\beta n)$.

Note that the time cost of fitness evaluation (for the entire population) normally dominates the one of selection and the larger the number of observations in the training set $n$, the less the cost of selection will matter.
We remark that steps \num{2} and \num{3} can be implemented in terms of $(\beta \times n)$-dimensional matrix operations, if desired (e.g., in our \texttt{python} implementation, we leverage \texttt{numpy}~\cite{van2011numpy}).

\paragraph{Linear scaling}\label{sec:adapting-linear-scaling}
We can easily include linear scaling when computing the fitness on all bootstrap samples.
Linear scaling is an efficient and effective method to improve the performance of GP in regression~\cite{keijzer2003improving,keijzer2004scaled}.
It consists of computing and applying two coefficients $a,b$ to perform an affine transformation of the output that optimally minimizes the training (optionally, root) mean squared error as in $\text{MSE}^{a,b}(y, o) = \frac{1}{n} \sum_{i=1}^n \left(y_i - (a + bo_i)\right)^2$.
These coefficients are:
\begin{equation}\label{eq:linear-scaling}
    a = \bar{y} - b \bar{o}, \ \ \ \ 
    b = \frac{ \sum_{i=1}^n \left( y_i - \bar{y} \right) \left( o_i - \bar{o} \right) } { \sum_{i=1}^n \left( o_i - \bar{o} \right)^2 },
\end{equation}
where $\bar{y}$ (resp., $\bar{o}$) denote the arithmetic mean of the label (resp., output) over the training set $\mathbb{T}$.
SotA GP algorithms often include linear scaling (or regression in some form)~\cite{cava2020learning,cava2018learning,virgolin2020improving,virgolin2019linear,vzegklitz2020benchmarking}.

We incorporate linear scaling in our approach by computing $\beta$ coefficients $a_j,b_j$ to scale each $\mathbb{T}_j$-specific output in a similar fashion to how step \num{3} of the previous section is performed.
This requires to add an $\mathcal{O}(\beta n)$ term to the left-hand side of \Cref{eq:fitness-cost}, which does not change the asymptotics.
Implementation can again rely on matrix operations for the sake of speed (see our code).

\section{Experimental setup}\label{sec:experimental-setup}
We attempt to (mostly) reproduce the experimental settings used in~\cite{rodrigues2020ensemble}, to which we compare in terms of classification. 
Specifically, we use $n_\text{pop}=\num{500}$, the selection method described in \Cref{sec:our-selection}, and variation by subtree crossover and subtree mutation with equal probability (\num{0.5}). 
We use the uniform random depth node selection method for variation~\cite{pawlak2014semantic} to oppose bloat.
If an offspring with more than \num{500} nodes is generated, we discard it and clone the parent.

We use ramped half-and-half initialization with tree heights \num{2}--\num{6}~\cite{poli2008field}.
The function set is $\{ +, -, \times, \widetilde{\div}, \widetilde{\sqrt{\cdot}}, \widetilde{\log} \}$, with the last three operators implementing protection by, respectively, $\widetilde{\div}(a,b) := a \times \text{sign}(b) / \left( |b| + \varepsilon \right)$, $\widetilde{\sqrt{x}} := \sqrt{ |x| }$, $\widetilde{\log}(x) := \log ( |x| + \varepsilon )$, with $\text{sign}(0) := 1$ and $\varepsilon = 10^{-10}$.
Alongside the features, we include an ephemeral random constant terminal~\cite{poli2008field} (even though \cite{rodrigues2020ensemble} chose not to) with values sampled from $\mathcal{U}(-5, 5)$, because ephemeral random constants can improve performance~\cite{poli2008field,virgolin2020improving} (and other algorithms we compare to use them). 
2SEGP needs only one additional hyper-parameter compared to classic GP, i.e., the desired ensemble size $\beta$.
We set $\beta = 0.1 \times n_\text{pop} = 50$ as a rule of thumb.
We analyze other settings of $\beta$ in \Cref{sec:insights-beta}.

We use z-score data standardization as advised in~\cite{dick2020feature}. 
We include linear scaling for both regression and classification tasks.
In our case it is plausible to apply linear scaling in classification (prior to rounding the output to the nearest class) since the considered problems are binary (the label is \num{0} or \num{1}). 
For completeness, we also include results without linear scaling for classification.

A run consists of \num{100} generations. 
We conduct \num{40} independent runs to account for the randomness of both GP and training-test splitting, for which we use a \SI{70}{\percent}--\SI{30}{\percent} ratio as in~\cite{rodrigues2020ensemble,vzegklitz2020benchmarking}.
Statistical significance is evaluated using pairwise non-parametric Mann-Whitney $U$ tests with $p\text{-value} < 0.05$ and Holm-Bonferroni correction~\cite{holm1979simple,mann1947test}.
In particular, we say that an algorithm is among the best ones if no other performs significantly better.

\section{Competing algorithms and considered datasets}\label{sec:competing}
\paragraph{Classification.}
For comparisons on classification problems, the first set of results we consider was provided by the authors of~\cite{rodrigues2020ensemble}.
From~\cite{rodrigues2020ensemble}, we report the results of the best-performing ensemble algorithm ``ensemble GP with weighted voting'' (eGPw); the best-performing non-ensemble algorithm ``Multidimensional Multiclass GP with Multidimensional Populations'' (M3GP), and classic GP (cGP).
M3GP in particular is a SotA GP-based feature construction approach.
In~\cite{rodrigues2020ensemble}, M3GP is found to outperform most of the other (GP and non-GP) algorithms, including random forest.

We further include our own re-implementation of ``Diverse Niching Genetic Programming'' (DivNichGP), made by following~\cite{wang2019novel}, and that we make available at \url{https://github.com/marcovirgolin/DivNichGP}.
For DivNichGP, we maintain equal subtree crossover and mutation probability, but also allow reproduction to take place with a $5\%$ rate, to follow the settings of the original paper.
DivNichGP internally uses tournament selection; We set this to size \num{8} (as for our cGP for regression, described below).
For DivNichGP's niching mechanism, we use the same distance threshold of \num{0} and maximal niche size of \num{1} as in~\cite{wang2019novel}.
Since DivNichGP uses a validation set to aggregate the ensemble, we build a pseudo-validation set by taking the out-of-bag observations of the last-sampled realization of the training set.
All the other settings are as in~\Cref{sec:experimental-setup}.

The datasets we consider for classification are the five real-world datasets used in~\cite{rodrigues2020ensemble} that are readily available from the UCI repository~\cite{dua2017uci}.
We refer to~\cite{rodrigues2020ensemble} for details on these datasets.

\paragraph{Regression.} 
For regression, we report results from~\cite{vzegklitz2020benchmarking} (see their Table 7), i.e., median test errors of SotA GP regression algorithms.
These algorithms are ``Evolutionary Feature Synthesis'' (EFS)~\cite{arnaldo2015building}, ``Genetic Programming Toolbox for The Identification of Physical Systems'' (GPTIPS)~\cite{searson2015gptips,searson2010gptips} (and a modified version mGPTIPS that uses settings similar to those of EFS), and ``Geometric Semantic Genetic Programming with Reduced Trees'' (GSGP-Red)~\cite{martins2018solving}.
We refer to~\cite{vzegklitz2020benchmarking} for the settings and choices made for these algorithms.

We further include a home-baked version of cGP that uses tournament selection of size \num{8} (we also experimented with size \num{4} and truncation selection, but they performed worse), with all other settings being as explained before. 
We use again our re-implementation of DivNichGP.
Next, as additional ensemble learning GP algorithm, we consider the ``Spatial Structure with Bootstrapped Elitism'' (SS+BE) algorithm proposed in~\cite{dick2018evolving}, by means of results that were provided by the first author of the work. 
The settings for SS+BE are slightly different from those of 2SEGP in that they follow those presented in~\cite{dick2018evolving}, as prescribed by the author:
Evolution runs for \num{250} generations, with a population of size $196$, and using a $14\times14$ toroidal grid distribution.

Next, we consider two recent algorithms that improve variation.
The first is the GP version of the Gene-pool Optimal Mixing Evolutionary Algorithm (GP-GOMEA)~\cite{virgolin2020improving,virgolin2017scalable}.
GP-GOMEA uses a form of crossover that preserves high-mutual information patterns.
Since GP-GOMEA requires relatively large population sizes to infer meaningful patterns but converges quickly, we shift resources between population size and number of generations, i.e., we set $n_\text{pop}=5000$ and use only \num{10} generations.
Moreover, GP-GOMEA uses a fixed tree template representation: We set the template height to \num{7} so that up to \num{255} nodes can be hosted (half the maximum allowed size for the other algorithms).
Second and last, we include the linear scaling-enhanced version of the semantic operator Random Desired Operator~\cite{pawlak2014semantic,wieloch2013running}, denoted by $\text{RDO}^{\times\text{LS}}_{+\text{LS}}$ in~\cite{virgolin2019linear}.
$\text{RDO}^{\times\text{LS}}_{+\text{LS}}$ uses a form of semantic-driven mutation based on the internal computations of GP subtrees and a library of pre-computed subtrees.
We use the traditional ``population-based'' library, updated every generation and storing up to \num{500} subtrees, up to \num{12} deep.

Like for 2SEGP, linear scaling (or some similar form thereof, see~\cite{vzegklitz2020benchmarking}) is also used for the other algorithms (except for GSGP-Red for which it was not used~\cite{vzegklitz2020benchmarking}).
We remark that while the generational cost of 2SEGP is only marginally larger than the one of cGP (as explained in \Cref{sec:method}), the same is often not true for the competing SotA algorithms, some of which take substantially more time to run (we refer to the respective papers for details). 
Hence, in many comparisons, 2SEGP can be considered to be disadvantaged.

For the sake of reproducibility we rely once more on datasets used in previous work, and this time specifically on the four real-world UCI datasets of~\cite{vzegklitz2020benchmarking}.

\section{Benchmark results}\label{sec:benchmarkingresults}

\paragraph{Classification}
\Cref{tab:comparison-rodrigues2020ensemble} shows the accuracy obtained by eGPw, M3GP, DivNichGP, cGP, and of course 2SEGP, the latter with and without linear scaling.
At training time, M3GP is significantly best on three out of five datasets, while 2SEGP is second-best.
Compared to eGPw and DivNichGP, which also evolve ensembles, 2SEGP performs better (on Heart significantly so), except for on Parks when linear scaling is disabled.
This is the only dataset where we observe a substantial drop in performance when linear scaling is turned off.
When testing, due to the generalization gap and the Holm-Bonferroni correction, less results are significantly different. 
This is evident for BCW.
Compared to M3GP, 2SEGP is significantly better on Parks, but worse on Sonar.
On Heart, M3GP is no longer superior, as substantial performance is lost when testing.
Note also that DivNichGP, possibly because it uses a (pseudo-)validation set to choose the final ensemble, exhibits slightly (but not significantly) better generalization than 2SEGP on Heart and Iono.
Overall, despite being simpler, 2SEGP fares well against DivNichGP, eGPw, and even M3GP.

\begin{table*}[]
    \centering
    \caption{
        Median accuracy (higher is better) $\pm$ interquartile range of 2SEGP, 2SEGP w/o linear scaling (w/oLS), DivNichGP, eGPw, M3GP, and cGP on the UCI datasets of~\cite{rodrigues2020ensemble}.
        Underlined results are best, i.e., not significantly worse than any other.
    }
    \label{tab:comparison-rodrigues2020ensemble}
    \setlength{\tabcolsep}{0.8\tabcolsep}
    \begin{tabular}{l 
    S[table-format=1.3]@{\hspace{0.05\tabcolsep}}l
    S[table-format=1.3]@{\hspace{0.05\tabcolsep}}l
    S[table-format=1.3]@{\hspace{0.05\tabcolsep}}l
    S[table-format=1.3]@{\hspace{0.05\tabcolsep}}l
    S[table-format=1.3]@{\hspace{0.05\tabcolsep}}l
    }
\toprule
        & \multicolumn{10}{c}{Training}\\ 
        \cmidrule(lr){2-11}
        Algorithm & \multicolumn{2}{c}{BCW} & \multicolumn{2}{c}{Heart} & \multicolumn{2}{c}{Iono} & \multicolumn{2}{c}{Parks} & \multicolumn{2}{c}{Sonar} \\
        \midrule
2SEGP (ours) & \underline{0.995} & \scriptsize{$\pm$0.005} & 0.944 & \scriptsize{$\pm$0.022} & \underline{0.976} & \scriptsize{$\pm$0.017} & 0.948 & \scriptsize{$\pm$0.011} & 0.966 & \scriptsize{$\pm$0.034} \\
w/oLS (ours) & 0.995 & \scriptsize{$\pm$0.006} & 0.947 & \scriptsize{$\pm$0.021} & 0.978 & \scriptsize{$\pm$0.012} & 0.892 & \scriptsize{$\pm$0.021} & 0.959 & \scriptsize{$\pm$0.036} \\
DNGP & 0.979 & \scriptsize{$\pm$0.010} & 0.915 & \scriptsize{$\pm$0.021} & 0.955 & \scriptsize{$\pm$0.015} & 0.931 & \scriptsize{$\pm$0.057} & 0.924 & \scriptsize{$\pm$0.043}\\
eGPw & 0.983 & \scriptsize{$\pm$0.008} & 0.907 & \scriptsize{$\pm$0.025} & 0.884 & \scriptsize{$\pm$0.032} & 0.923 & \scriptsize{$\pm$0.042} & 0.924 & \scriptsize{$\pm$0.034}\\
M3GP & 0.971 & \scriptsize{$\pm$0.002} & \underline{0.970} & \scriptsize{$\pm$0.017} & 0.932 & \scriptsize{$\pm$0.042} & \underline{0.981} & \scriptsize{$\pm$0.024} & \underline{1.000} & \scriptsize{$\pm$0.012} \\
cGP & 0.964 & \scriptsize{$\pm$0.016} & 0.825 & \scriptsize{$\pm$0.033} & 0.773 & \scriptsize{$\pm$0.060} & 0.842 & \scriptsize{$\pm$0.077} & 0.769 & \scriptsize{$\pm$0.055} \\
\midrule
& \multicolumn{10}{c}{Test}\\ 
        \cmidrule(lr){2-11}
        Algorithm & \multicolumn{2}{c}{BCW} & \multicolumn{2}{c}{Heart} & \multicolumn{2}{c}{Iono} & \multicolumn{2}{c}{Parks} & \multicolumn{2}{c}{Sonar} \\
        \midrule
2SEGP (ours) & \underline{0.965} & \scriptsize{$\pm$0.018} & \underline{0.815} & \scriptsize{$\pm$0.062} & \underline{0.896} & \scriptsize{$\pm$0.047} & \underline{0.936} & \scriptsize{$\pm$0.012} & 0.738 & \scriptsize{$\pm$0.067} \\
w/oLS (ours) & \underline{0.965} & \scriptsize{$\pm$0.013} & 0.809 & \scriptsize{$\pm$0.049} & 0.896 & \scriptsize{$\pm$0.047} & 0.885 & \scriptsize{$\pm$0.031} & 0.754 & \scriptsize{$\pm$0.067} \\
DNGP & \underline{0.959} & \scriptsize{$\pm$0.019} & \underline{0.815} & \scriptsize{$\pm$0.049} & \underline{0.901} & \scriptsize{$\pm$0.026} & 0.917 & \scriptsize{$\pm$0.055} & 0.730 & \scriptsize{$\pm$0.063} \\
eGPw & \underline{0.956} & \scriptsize{$\pm$0.018} & \underline{0.790} & \scriptsize{$\pm$0.034} & 0.830 & \scriptsize{$\pm$0.057} & 0.822 & \scriptsize{$\pm$0.064} & 0.762 & \scriptsize{$\pm$0.060} \\
M3GP & \underline{0.957} & \scriptsize{$\pm$0.014} & 0.778 & \scriptsize{$\pm$0.069} & \underline{0.871} & \scriptsize{$\pm$0.057} & 0.897 & \scriptsize{$\pm$0.051} & \underline{0.810} & \scriptsize{$\pm$0.071} \\
cGP & \underline{0.961} & \scriptsize{$\pm$0.018} & 0.784 & \scriptsize{$\pm$0.049} & 0.745 & \scriptsize{$\pm$0.057} & 0.797 & \scriptsize{$\pm$0.102} & 0.714 & \scriptsize{$\pm$0.044}\\
        \bottomrule    
    \end{tabular}
\end{table*}

\paragraph{Regression}
\Cref{tab:comparison-vzegklitz2020benchmarking} shows the results of 2SEGP, DivNichGP, SS+BE, GP-GOMEA, $\text{RDO}^{\times\text{LS}}_{+\text{LS}}$, cGP, and the algorithms from~\cite{vzegklitz2020benchmarking} (only test is reported in their Table~\num{7}) in terms of Root Mean Squared Error (RMSE).
2SEGP always outperforms DivNichGP with the exception of training on ENH and testing on ENC.
Similarly, 2SEGP outperforms SS+BE on almost all cases (not when testing on ENC).
2SEGP is also competitive with the SotA algorithms, as it is only significantly worse than GP-GOMEA and $\text{RDO}^{\times\text{LS}}_{+\text{LS}}$ on ENH when testing.
On ASN, 2SEGP is not matched by any other algorithm.
Interestingly, our implementation of cGP achieves rather good results on most datasets, and performs better in terms of median RMSE than some of the SotA algorithms from~\cite{vzegklitz2020benchmarking}.

\begin{table*}[]
    \centering
    \caption{
        Median RMSE (smaller is better) $\pm$ interquartile range of the considered algorithms on the UCI datasets of~\cite{vzegklitz2020benchmarking}. 
        Median results of GPTIPS, mGPTIPS, EFS, and GSGP-Red are reported from~\cite{vzegklitz2020benchmarking}.
        Underlined results are best, i.e., not significantly worse than any other (excl.~the algs.~from~\cite{vzegklitz2020benchmarking} as we only have medians).
        Best median-only test results are starred.
    }
    \label{tab:comparison-vzegklitz2020benchmarking}
    \begin{tabular}{l 
    S[table-format=1.3]@{\hspace{0.05\tabcolsep}}l
    S[table-format=1.3]@{\hspace{0.05\tabcolsep}}l
    S[table-format=1.3]@{\hspace{0.05\tabcolsep}}l
    S[table-format=1.3]@{\hspace{0.05\tabcolsep}}l
    }
    \toprule
        & \multicolumn{8}{c}{Training} \\ 
        \cmidrule(lr){2-9}
        Algorithm & \multicolumn{2}{c}{ASN} & \multicolumn{2}{c}{CCS} & \multicolumn{2}{c}{ENC} & \multicolumn{2}{c}{ENH} \\
        \midrule
2SEGP (ours) & \underline{2.899} & \scriptsize{$\pm$0.290} & \underline{5.822} & \scriptsize{$\pm$0.353} & \underline{1.606} & \scriptsize{$\pm$0.200} & \underline{0.886} & \scriptsize{$\pm$0.556}\\
DivNichGP & 3.360 & \scriptsize{$\pm$0.343} & 6.615 & \scriptsize{$\pm$0.454} & 1.809 & \scriptsize{$\pm$0.190} & \underline{1.079} & \scriptsize{$\pm$0.415}  \\
SS+BE & 3.271 & \scriptsize{$\pm$0.316} & 6.517 & \scriptsize{$\pm$0.412} & 1.882 & \scriptsize{$\pm$0.363} & 1.190 & \scriptsize{$\pm$0.291} \\
GP-GOMEA & 3.264 & \scriptsize{$\pm$0.172} & 6.286 & \scriptsize{$\pm$0.300} & \underline{1.589} & \scriptsize{$\pm$0.079} & \underline{0.739} & \scriptsize{$\pm$0.138}  \\
$\text{RDO}^\text{xLS}_\text{+LS}$ & 3.482 & \scriptsize{$\pm$0.172} & 6.476 & \scriptsize{$\pm$0.249} & 1.703 & \scriptsize{$\pm$0.125} & \underline{0.819} & \scriptsize{$\pm$0.186} \\
cGP & 3.160 & \scriptsize{$\pm$0.295} & 6.279 & \scriptsize{$\pm$0.305} & 1.851 & \scriptsize{$\pm$0.441} & 1.196 & \scriptsize{$\pm$0.431} \\
\midrule
        & \multicolumn{8}{c}{Test} \\ 
        \cmidrule(lr){2-9} 
        Algorithm & \multicolumn{2}{c}{ASN} & \multicolumn{2}{c}{CCS} & \multicolumn{2}{c}{ENC} & \multicolumn{2}{c}{ENH} \\
        \midrule
2SEGP (ours) &  \underline{3.082}$^\star$ & \scriptsize{$\pm$0.438} & \underline{6.565} & \scriptsize{$\pm$0.439} & \underline{1.801} & \scriptsize{$\pm$0.263} & 0.961 & \scriptsize{$\pm$0.553} \\
DivNichGP & 3.458 & \scriptsize{$\pm$0.487} & 7.031 & \scriptsize{$\pm$0.370} & \underline{1.930} & \scriptsize{$\pm$0.156} & 1.158 & \scriptsize{$\pm$0.398} \\
SS+BE & 3.416 & \scriptsize{$\pm$0.333} & 6.986 & \scriptsize{$\pm$0.744} & 1.946 & \scriptsize{$\pm$0.380} & 1.204 & \scriptsize{$\pm$0.366} \\
GP-GOMEA & 3.346 & \scriptsize{$\pm$0.238} & \underline{6.777} & \scriptsize{$\pm$0.313} & \underline{1.702} & \scriptsize{$\pm$0.200} & \underline{0.804} & \scriptsize{$\pm$0.184} \\
$\text{RDO}^\text{xLS}_\text{+LS}$ & 3.579 & \scriptsize{$\pm$0.245} & 6.800 & \scriptsize{$\pm$0.423} & \underline{1.791} & \scriptsize{$\pm$0.180} & \underline{0.881} & \scriptsize{$\pm$0.309} \\
cGP & 3.359 & \scriptsize{$\pm$0.379} & \underline{6.759} & \scriptsize{$\pm$0.623} & \underline{2.041} & \scriptsize{$\pm$0.383} & 1.267 & \scriptsize{$\pm$0.556} \\
\hline
GPTIPS & 4.138 &  & 8.762 &  & 2.907 &  & 2.538 &  \\
mGPTIPS & 4.003 &  & 7.178 &  & 2.278 &  & 1.717 &  \\
EFS  & 3.623 &  & 6.429$^\star$ &  & 1.640$^\star$ &  & 0.546$^\star$ &  \\
GSGP-Red & 12.140 &  & 8.798 &  & 3.172 &  & 2.726 &  \\
        \bottomrule
    \end{tabular}
\end{table*}

\section{Insights}\label{sec:insights}
In this section, we provide insights about our proposal.
We begin by assessing the role of $\beta$ in terms of prediction error and time, including when the ensemble is formed by an SIEL-App.
Next, we investigate our selection method by ablation. 
Last but not least, we peek into the effect of classic GP variation in 2SEGP.
From now on, we consider the regression datasets.

\subsection{On the role of the ensemble size $\beta$}\label{sec:insights-beta}
We assess the performance gain (or loss) of the approach when $\beta$ is increased while the population size $n_\text{pop}$ is kept fixed.
We include a comparison to obtaining an ensemble by running independent cGP evolutions, i.e., as in a classic SIEL-App.
For 2SEGP, we scale $\beta$ (approx.) exponentially, i.e., $\beta=5$, $25$, $50$, $100$, $250$, $500$. 
For our SIEL-App, we use $\beta=1, 2, \dots, 10$, as running times of sequential executions quickly become prohibitive.
We include cGP, DivNichGP, and SS+BE in the comparison. 
All settings are as before (\Cref{sec:experimental-setup}).

\begin{figure}[h]
    \centering
    \includegraphics[width=0.8\linewidth]{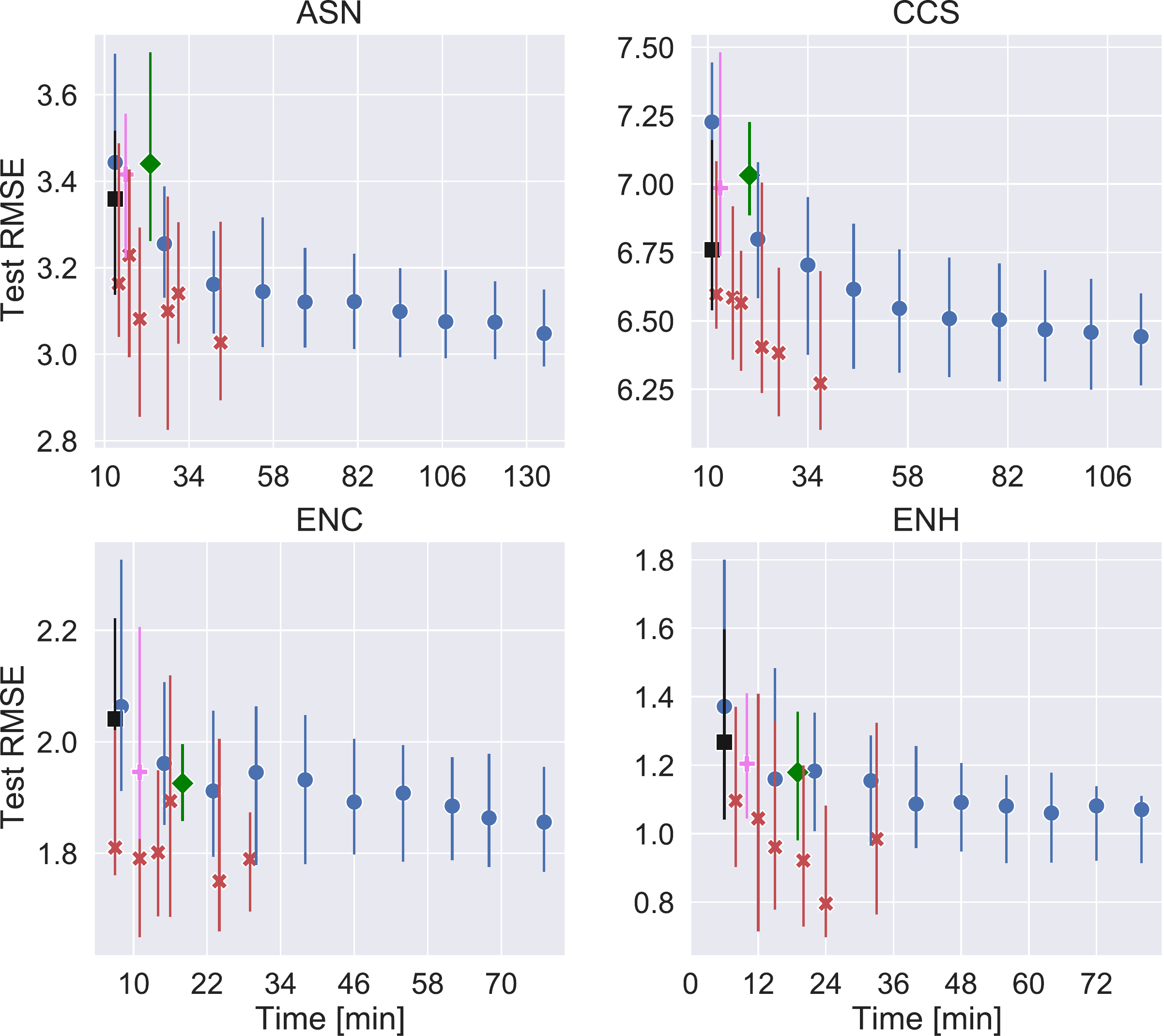}
    \caption{Distribution of test RMSE (median and interquartile range) w.r.t. average time taken by 2SEGP (in {\color{BrickRed}\textbf{red}}), our SIEL-App (in {\color{MidnightBlue}\textbf{blue}}), DivNichGP (in {\color{OliveGreen}\textbf{green}}), and SS+BE (in {\color{Rhodamine}\textbf{pink}}); or a single estimator by cGP (in \textbf{black}). 
    For 2SEGP, $\beta$ is scaled approximately exponentially (from left to right, $\beta$ is $5, 25, 50, 100, 250, 500$). 
    For our SIEL-App, $\beta$ is scaled linearly (from left to right, $\beta$ is $1, 2, 3, \dots, 10$).}
    \label{fig:role-of-beta}
\end{figure}

\paragraph{Role of $\beta$ in 2SEGP}
\Cref{fig:role-of-beta} shows the distribution of test RMSE against the average time taken when using different $\beta$ settings (results on the training set are not shown here but follow the same trends).  
For now we focus on 2SEGP (red crosses), and will consider the other algorithms later.
Larger ensembles seem to perform similarly to, or slightly better than, smaller ensembles, with diminishing returns.
Statistical tests between pairwise configurations of $\beta$ for 2SEGP reveal that most test RMSE distributions are not significantly different from each other ($p\text{-value} \geq 0.05$ except, e.g., between $\beta=5$ and $\beta=500$ on CCS; and between $\beta=5$ and $\beta=250$ on ENH).
In particular, we cannot refute the null hypothesis that larger $\beta$ leads to better performance because inter-run performance variability is relatively large (this is in part due to performance loss when testing).
Hence, setting $\beta$ to large values such as $\beta=1.0 \times n_\text{pop}$ results in a time cost increase for no marked performance gain.

\begin{figure}
    \centering
    \includegraphics[width=0.8\linewidth]{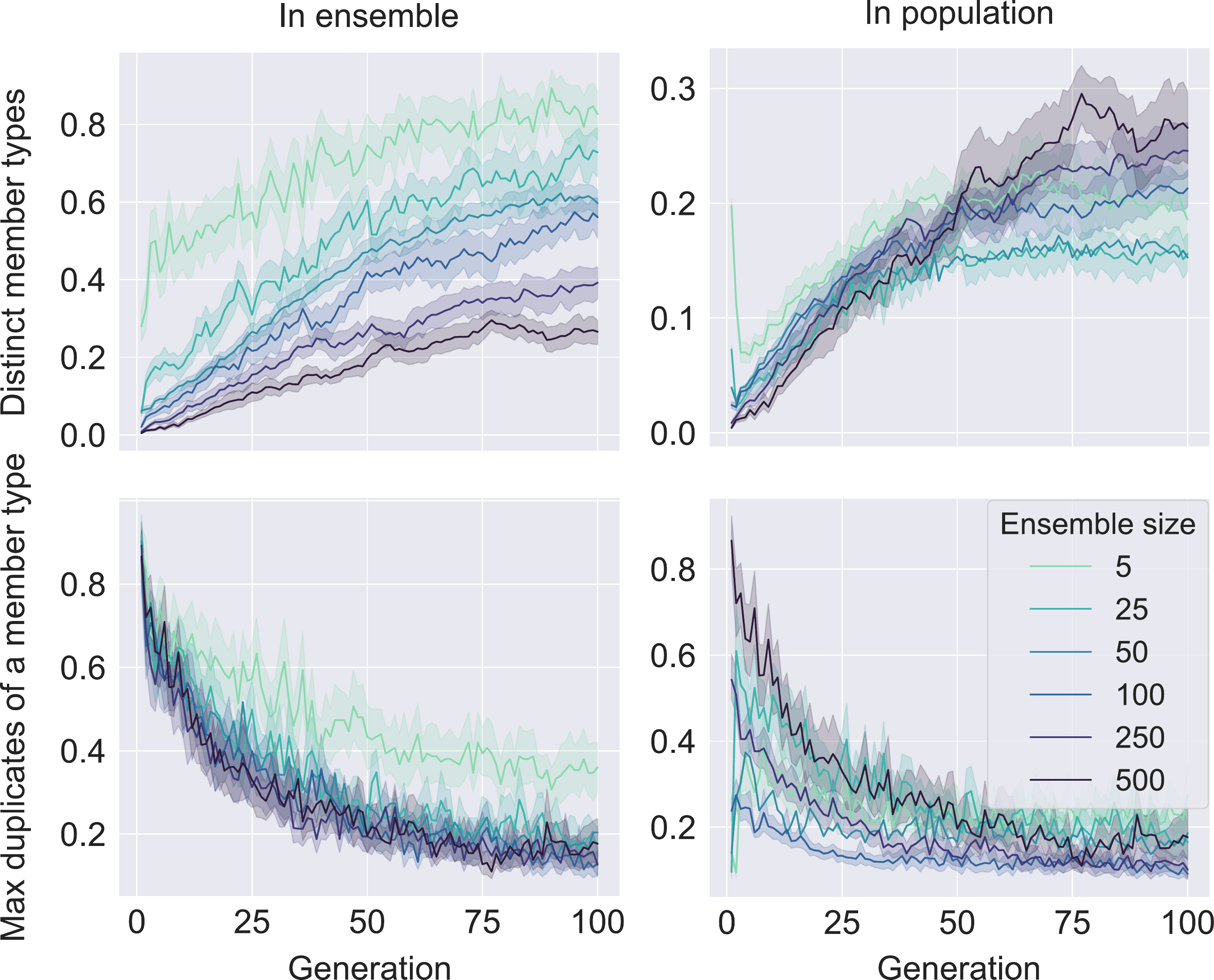}
    \caption{Mean and $95\%$ conf. intervals of $40$ runs on ASN (w. $n_\text{pop}=500$) of two aspects of diversity (top and bottom) as ratios over ensemble (left) or population size (right).}
    \label{fig:beta-ensemble-diversity}
\end{figure}

Delving deeper, \Cref{fig:beta-ensemble-diversity} shows, for different $\beta$ settings in ASN runs, how many individuals are different from one another during the evolution (with $n_\text{pop}=500$).
This is shown for the ensemble, i.e., the collection of $\beta$ individuals that are elite according to a $\mathbb{T}_j$-specific fitness value, and for the population.
We consider \emph{exact syntactical copies} (rather than, e.g., on semantic equivalence) to better assess the influence of selection, which copies individuals as they are.
The plots on the left show that, no matter how big $\beta$ is, only a very small number of distinct individuals are top-ranking across all the bootstrap samples at initialization. 
As the evolution proceeds, the larger $\beta$ is, the more the ensemble will be redundant.
The bottom-left plot shows that, no matter what $\beta$ is (excl. $\beta=5$), one-fifth of the final ensemble is made by duplicates of one type of individual.

The plots on the right show how $\beta$ affects the population.
We know that, in classic GP, duplicates can rapidly increase in early stages to then decrease later, when small modifications of a same root structure are generated~\cite{burke2004diversity,burlacu2013visualization,langdon2017long}.
This effect can be seen for small $\beta$ values. 
For (too) large $\beta$ values, a single generation is sufficient to annihilate population diversity. 
This is because our selection causes top-ranking individuals across the $\mathbb{T}_j$s to get $\beta$ copies, and at initialization only a few individuals have decent performance.
Nevertheless, considering \Cref{fig:role-of-beta}, this does not seem to break the algorithm in terms of test RMSE.
This could be explained by the fact that larger $\beta$ values also allow for larger diversity gains later on, as visible in the last generations of the top-right plot.
In fact, for large $\beta$ there are many $\mathbb{T}_j$s and thus a larger number of elites is maintained. 
Conversely, when $\beta$ is smaller (e.g., $5$ or $25$), less elites are possible and population diversification caps sooner.

Since many ensemble members can be duplicates, \emph{we can prune the ensemble obtained at the end of the run}. 
In fact, we remark that if one (takes a weighted average of the linear scaling coefficients shared by duplicate individuals and) removes duplicates, the ensemble retains the same predictive power. 
Pruning for $\beta=50$ already results in considerable reductions of the ensemble size, between $34\%$ (for ENH) and $75\%$ (for CCS) of the original size.

Overall, these results show that: (i) Performance-wise, 2SEGP is relatively robust to the setting of $\beta$; (ii) The ensemble may contain duplicates, but this does not represent an issue because duplicates can be trimmed off without any decrease of predictive power; and, ultimately, (iii) It is sensible to use values of $\beta$ between $5\%$ and $30\%$ of the population size.

\paragraph{Comparing with the SIEL-App.}
The time cost taken by our SIEL-App to form an ensemble of size $\beta$ is approximately $\beta$ times the time of performing a single cGP evolution, as expected (we address potential for parallelism in the last paragraph of this section).
As can be seen in \Cref{fig:role-of-beta}, 2SEGP can build larger ensembles in a fraction of the time taken by the SIEL-App, in line with our expectation from~\Cref{eq:fitness-cost}.
We also report the performance of SS+BE (run on a different machine by the first author of~\cite{dick2018evolving}) and DivNichGP for $\beta=50$. In brief, 2SEGP with reasonable settings (e.g., $\beta=25,50$) has a running time which is in the same ballpark of the times taken by SS+BE and DivNichGP, hence it is similarly efficient.

We now focus on comparing 2SEGP with the SIEL-App and start by considering the setting $\beta=5$ for both, i.e., in each plot, the first red point and the fifth blue point, respectively.
Interestingly, while 2SEGP uses only a fraction of the computational resources required to learn the ensemble compared to our SIEL-App, the ensembles obtained by the SIEL-App do not outperform the ensembles obtained by 2SEGP.
The SIEL-App starts to perform slightly better than cGP already with $\beta=2$, but at the cost of twice the running time.
Within that time, 2SEGP can use \num{50} bootstrap samples (3rd red dot) and typically obtains better performance than any other algorithm.
In general, given a same time limit, \emph{under-performing runs of 2SEGP are often better than or similar to average-performing runs of the SIEL-App}, thanks to the former being capable of evolving larger ensembles.
A downside of 2SEGP is that it obtains larger inter-run performance variance than the SIEL-App.
Nevertheless, this is only natural because the latter uses a new population to evolve each ensemble member.

We remark that if the population size (which we now denote by $|\mathbb{P}|$ for readability) and/or the number of generations ($G$) required by our SIEL-App are reduced as to make the SIEL-App match the computational expensiveness of 2SEGP, then the SIEL-App performs poorly.
This can be expected because (cfr. \Cref{sec:fitness-eval}):
\begin{equation}\label{eq:evolution-cost-comparison}
\begin{aligned}
 \text{Time cost of the SIEL-App} &\simeq \text{Time cost of 2SEGP}\\
 \beta G^\text{SIEL-App} |\mathbb{P}|^\text{SIEL-App} \ell n &\simeq
 G^\text{2SEGP} |\mathbb{P}|^\text{2SEGP}  n(\beta + \ell)\\
 G^\text{SIEL-App} |\mathbb{P}|^\text{SIEL-App} &\simeq 
 G^\text{2SEGP} |\mathbb{P}|^\text{2SEGP} \frac{\beta + \ell}{\beta \ell}.
\end{aligned}
\end{equation}
For example, if we assume $\ell=100$, set $\beta=50$, $G^\text{2SEGP}=100$, and $|\mathbb{P}|^\text{2SEGP}=500$, then a possible setting for the SIEL-App is $|\mathbb{P}|^\text{SIEL-App}=100$ and $G^\text{SIEL-App}=15$ (or vice versa); If we use the same settings but reduce the ensemble size to $\beta=5$, then for the SIEL-App we have $|\mathbb{P}|^\text{SIEL-App}=105$ and $G^\text{SIEL-App}=100$ (or vice versa).
With the former setting, we found that the SIEL-App cannot produce competitive results.
With the latter setting, the SIEL-App performed better, but still significantly worse than 2SEGP and cGP on all four regression datasets.

Finally, we must consider that, when an SIEL-App is used, each ensemble member can be evolved in parallel. 
If, e.g., $k \beta$ computation units are available, one can evolve a $\beta$-sized ensemble using $\beta$ parallel evolutions, each one parallelized on $k$ units.
Nevertheless, with 2SEGP, resources for parallelism can be fully invested into one population, which can consequently be increased in size if desired.
In other words, the results shown in this section regarding performance vs. time cost could in principle be rephrased in terms of performance vs. memory cost.
We leave an analysis of how an SIEL-App and 2SEGP compare in terms of the interplay between population size and parallel compute to future work.

\subsection{Ablation of selection}\label{sec:insight-selection}
We now investigate whether there is merit in partitioning the population during selection, as proposed in \Cref{sec:our-selection}.
If partitioning is disabled, one can no longer copy top-ranking estimators according to each $\mathbb{T}_j$.
We consider the following alternatives:
\begin{enumerate*}
    \item Survival according to truncation (Trunc) or tournament (Tourn) selection, based on the best fitness value among any $\mathbb{T}_j$---We call this strategy ``Push further What is Best'' (PWB);
    \item Like the previous point, but according to worse fitness value among any $\mathbb{T}_j$---We call this strategy ``Push What Lacks behind'' (PWL).
\end{enumerate*}
Note that also in~\cite{wen2016learning} individuals are ranked according to a PWB strategy (although the fitness values do not come from bootstrap samples).

We use the same settings of \Cref{sec:experimental-setup} (incl.~$\beta=0.1 \times n_\text{pop}$).
\Cref{tab:comparison-ablation-selection} shows test RMSEs obtained using our selection method and the ablated versions. 
It can be noted that the ablated versions perform worse than our selection method, with a few exceptions for tournament selection with size \num{8} on ENC or ENH.
In fact, the performance of tournament selection is the closest to the one of our selection. 
Using PWB or PWL leads to mixed results across the datasets, except when tournament selection with size \num{8} is used, where PWL is always better in terms of median results. 
Still, the proposed selection method leads to either equal or better performance.

\begin{table}[]
    \centering
    \caption{
        Median test RMSE $\pm$ interquartile range of our selection method and its ablations.
        Tournament size is \num{4} or \num{8}.
        Underlined results are best (not sig.~worse than any other).
    }
    \label{tab:comparison-ablation-selection}
    \begin{tabular}{l
    S[table-format=1.3]@{\hspace{0.05\tabcolsep}}l
    S[table-format=1.3]@{\hspace{0.05\tabcolsep}}l
    S[table-format=1.3]@{\hspace{0.05\tabcolsep}}l
    S[table-format=1.3]@{\hspace{0.05\tabcolsep}}l
    }
    \toprule
        Selection & \multicolumn{2}{c}{ASN} & \multicolumn{2}{c}{CCS} & \multicolumn{2}{c}{ENC} & \multicolumn{2}{c}{ENH} \\
        \midrule
        Ours & \underline{3.082} & \scriptsize{$\pm$0.438} & \underline{6.565} & \scriptsize{$\pm$0.439} & \underline{1.801} & \scriptsize{$\pm$0.263} & \underline{0.961} & \scriptsize{$\pm$0.553} \\
        
        {Trunc\textsuperscript{PWB}} & 3.727 & \scriptsize{$\pm$0.292} & 7.347 & \scriptsize{$\pm$0.489} & 2.187 & \scriptsize{$\pm$0.311} & 1.593 & \scriptsize{$\pm$0.449} \\
        
        {Trunc\textsuperscript{PWL}} & 3.689 & \scriptsize{$\pm$0.310} & 7.373 & \scriptsize{$\pm$0.468} & 2.154 & \scriptsize{$\pm$0.242} & 1.605 & \scriptsize{$\pm$0.310} \\
        
        {$\text{Tourn}^\text{PWB}_\text{4}$} & 3.527 & \scriptsize{$\pm$0.372}  & 6.996 & \scriptsize{$\pm$0.439} & 1.977 & \scriptsize{$\pm$0.479}  & 1.299 & \scriptsize{$\pm$0.302} \\ 
        
        {$\text{Tourn}^\text{PWL}_\text{4}$} & 3.569 & \scriptsize{$\pm$0.517} & 7.025 & \scriptsize{$\pm$0.443} & 1.946 & \scriptsize{$\pm$0.267} & 1.314& \scriptsize{$\pm$0.402}  \\ 
        
        {$\text{Tourn}^\text{PWB}_\text{8}$} & 3.440 & \scriptsize{$\pm$0.485} & 7.042 & \scriptsize{$\pm$0.475} & 1.938 & \scriptsize{$\pm$0.361} & \underline{1.137} & \scriptsize{$\pm$0.427} \\
        
        {$\text{Tourn}^\text{PWL}_\text{8}$} & 3.371 & \scriptsize{$\pm$0.338} & 6.896 & \scriptsize{$\pm$0.541} & \underline{1.876} & \scriptsize{$\pm$0.189} & \underline{1.023} & \scriptsize{$\pm$0.370}  \\
        \bottomrule
    \end{tabular}
\end{table}

\begin{figure}[h]
    \centering
    \includegraphics[width=\linewidth]{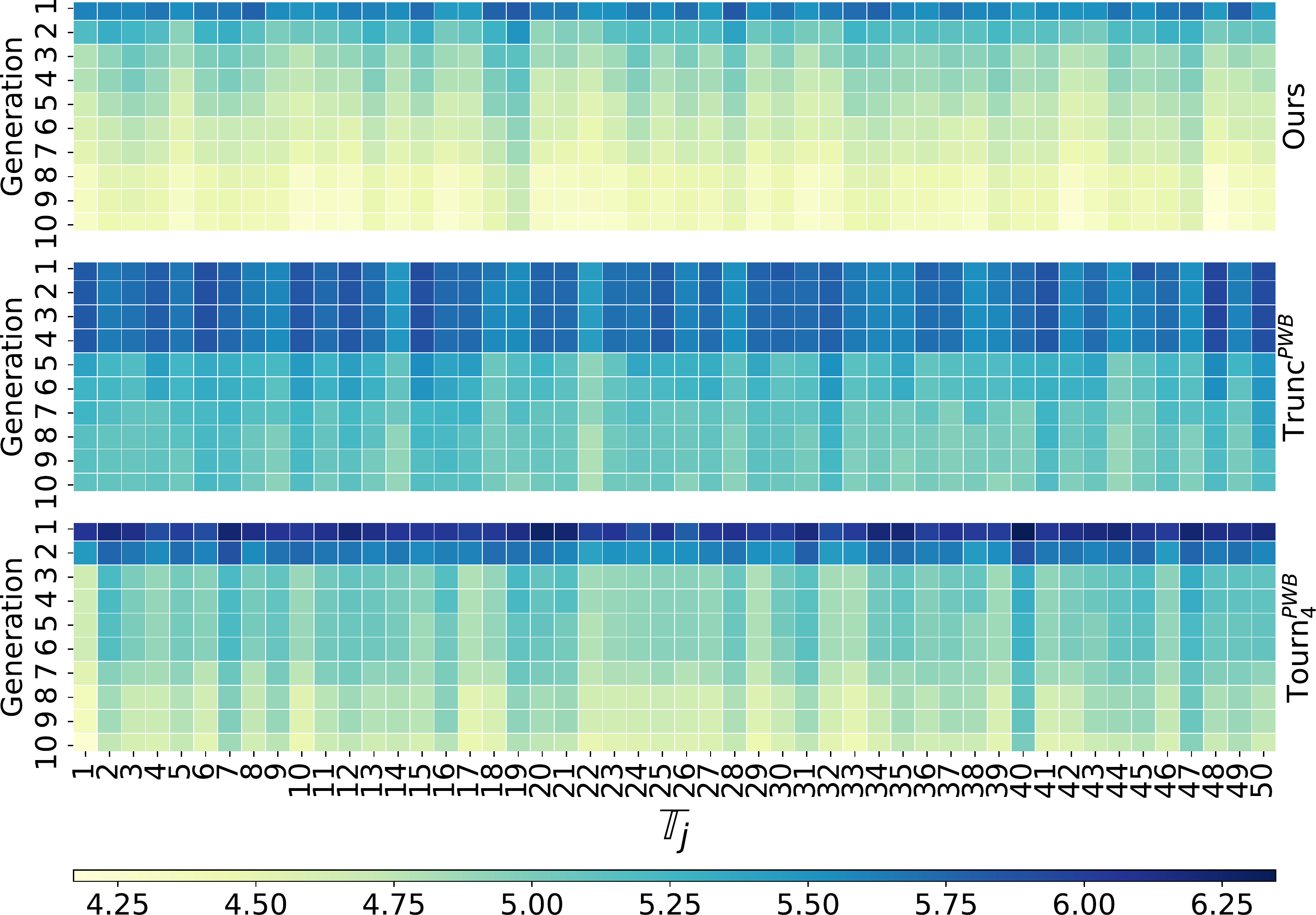}
    \caption{
        Training RMSEs of the best-found estimators for each $\mathbb{T}_j$ across \num{10} generations on ASN (lighter is better).
    }
    \label{fig:ensemble-fitness-selection-ablation}
\end{figure}

\Cref{fig:ensemble-fitness-selection-ablation} shows how the fitness values of the ensemble evolve using our selection method and the two PWB ablated versions, for one random run on ASN (we do not show the average of multiple runs as run-specific trends cancel out).
It can be seen that ablated truncation performs worse than the other two, and that our selection leads to the smallest RMSEs.
At the same time, our selection leads to rather uniform decrease of best-found RMSEs across the bootstrap samples.
Conversely, when using $\text{Tourn}^\text{PWB}_\text{4}$, some RMSEs remain large compared to the rest, e.g., notably so for $\mathbb{T}_7$, $\mathbb{T}_{40}$, and $\mathbb{T}_{47}$.

These results indicate that it is important to include partitioning as to promote uniform improvement across the bootstrap samples.
Since tournament selection performs rather well, and in particular better than simple truncation selection, it would be worth studying whether our selection method can be improved by incorporating tournaments in place of truncations, or SotA selection methods such as $\varepsilon$-lexicase selection~\cite{la2019probabilistic,la2016epsilon}.

\begin{figure}
    \centering
    \includegraphics[width=\linewidth]{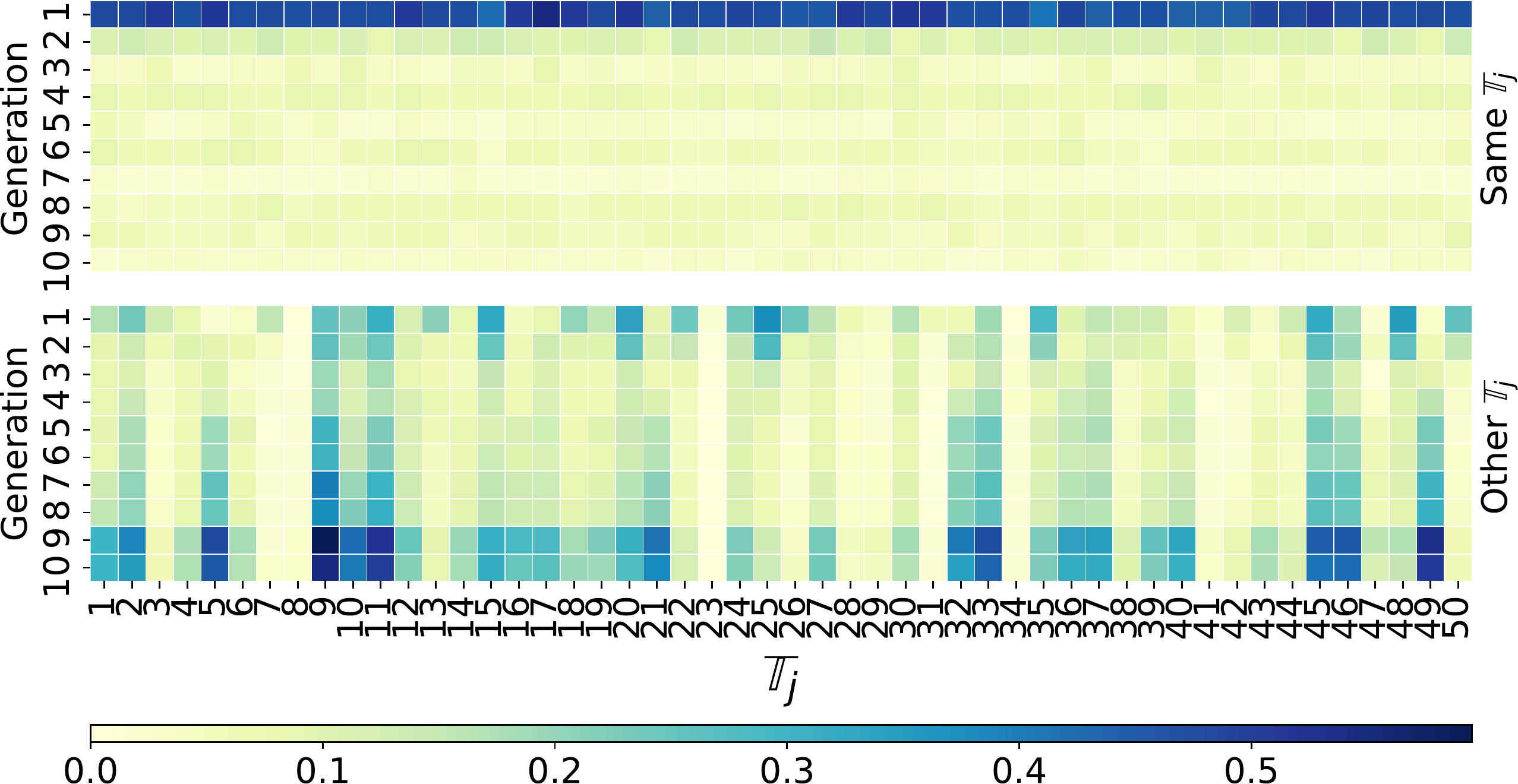}
    \caption{
        Frequency of producing offspring with smaller RMSE than their parents for the first \num{10} generations of a random run on ASN (darker is better).
    }
    \label{fig:rate-improvement-variation}
\end{figure}

\subsection{Evolvability by classic variation}\label{sec:improvements-variation}
In our experiments, we relied on classic subtree crossover and subtree mutation.
Our intuition was that mating between different individuals would be beneficial even if they rank better according to different bootstrap samples. 
To understand whether the use of classic variation is sufficient, we look at evolvability~\cite{verel2003bottlenecks}, here expressed as the frequency by which variation produces offspring that are fitter than their parents. 
We consider two aspects:
\begin{enumerate*}
    \item \emph{Same-$\mathbb{T}_j$ improvement}: Frequency of producing an offspring that has a better $j$th fitness value than the parent;
    \item \emph{Other-$\mathbb{T}_j$ improvement}: Frequency of producing an offspring that has an equal or worse $j$th fitness value than the parent, but better $k\text{th} \neq j\text{th}$ fitness value than the parent.
\end{enumerate*}

\Cref{fig:rate-improvement-variation} shows the ratios of improvement for the first \num{10} generations of a random run on ASN.
Not only Other-$\mathbb{T}_j$ improvements are frequent, they can be more frequent than Same-$\mathbb{T}_j$ improvements (we observe the same in other runs). 
So, an unsuccessful variation event w.r.t. one bootstrap sample can actually be successful w.r.t. to another bootstrap sample (see, e.g., the column for $\mathbb{T}_{49}$).
Thus, classic variation is already able to make the population improve across different realizations of the training set.
This corroborates our proposal of leaving classic variation untouched for the sake of simplicity.
Nevertheless, improvements may be possible by incorporating (orthogonal) SotA variation methods~\cite{moraglio2012geometric,pawlak2014semantic,virgolin2020improving}, or strategies for restricted mating and speciation~\cite{fonseca1995multiobjective,luong2014multi,stanley2002evolving}.

\section{Conclusions and future work}\label{sec:conclusion}
We show that small changes are sufficient to make an otherwise-classic GP algorithm evolve bagging ensembles efficiently and effectively.
Efficiency is a consequence of requiring only a single evolution over a single population where the nature of bootstrap sampling is exploited to perform fast fitness evaluations across all realizations of the training set. 
Effectiveness is perhaps somewhat surprising: The proposed algorithm can often match or even outperform state-of-the-art GP algorithms, despite being much simpler.
In light of these results, we argue that GP can be considered to be naturally suited to evolve bagging ensembles, which come (almost) for free in terms of computation cost.

There are a number of avenues for future work worth exploring.
Perhaps a first step could consist of studying whether it is possible to decouple selection pressure from the number of bootstrap samples. This would improve diversity preservation at the early stages of the evolution and possibly ultimately enhance ensemble quality, especially when one wishes to use a small population. 
Next, it will be interesting to integrate methods proposed in complex GP algorithms that are orthogonal and complementary to our approach, such as novel variation and ensemble aggregation methods. Designing ``ensemble-friendly'' versions of state-of-the-art selection methods (e.g., $\varepsilon$-lexicase selection~\cite{la2016epsilon}) could also be very beneficial, and porting knowledge from ensemble learning algorithms of different nature could lead to further improvements~\cite{probst2017tune,probst2019hyperparameters}.
Importantly, it would be natural to explore whether the fitness evaluation and selection changes proposed here can be applied to other types of evolutionary algorithms, e.g., to efficiently learn ensembles when optimizing the parameters or the topology of neural networks~\cite{stanley2002evolving}.
Last but not least, we remark that by learning an ensemble of many estimators, one loses an advantage of GP: The possibility to interpret the final solution~\cite{cano2013interpretable,guidotti2018survey,lensen2020genetic,virgolin2020explaining,virgolin2020learning}.
Nevertheless, future work could explore integrating ensemble methods for feature importance and prediction confidence estimation~\cite{dick2017sensitivity,kotanchek2008trustable,lakshminarayanan2017simple}, which are other relevant aspects to trust machine learning.

\section*{Acknowledgments}
I am deeply thankful to Nuno M. Rodrigues, Jo{\~a}o E. Batista, and Sara Silva from University of Lisbon, Lisbon, Portugal, for sharing the complete set of results of~\cite{rodrigues2020ensemble}, and to Grant Dick from University of Otago, Dunedin, New Zeland, for providing results for SS+BE~\cite{dick2018evolving}.
I thank Grant again, alongside Eric Medvet from University of Trieste, Trieste, Italy, and Mattias Wahde from Chalmers University of Technology, Gothenburg, Sweden, for the insightful discussions and feedback that helped improving this paper.
Part of this work was carried out on the Dutch national e-infrastructure with the support of SURF Cooperative.

\bibliographystyle{splncs04}
\bibliography{bib}

\end{document}